\documentclass{article}

\usepackage[final]{airov24-arxiv}

\usepackage[utf8]{inputenc} 
\usepackage[T1]{fontenc}    
\usepackage[]{hyperref}     
\usepackage{url}            
\usepackage{booktabs}       
\usepackage{amsmath}        
\usepackage{amssymb}
\usepackage{amsfonts}       
\usepackage{nicefrac}       
\usepackage{microtype}      
\usepackage{xcolor}         
\usepackage{graphicx}
\usepackage{listings}
\usepackage{tabularx}
\usepackage{multirow}

\newcolumntype{C}{>{\centering\arraybackslash}X}
\newcolumntype{R}{>{\raggedleft\arraybackslash}X}

\definecolor{commentcolor}{rgb}{0,0.6,0}
\definecolor{stringcolor}{rgb}{0.58,0,0.82}

\title{Flexible and Efficient Surrogate Gradient Modeling with Forward Gradient Injection}

\author{%
  Sebastian Otte\\
  Adaptive AI Lab\\
  Institute of Robotics and Cognitive Systems\\
  University of Lübeck, Germany\\
  \texttt{sebastian.otte@uni-luebeck.de}
}

\begin{document}

\maketitle

\begin{abstract}
Automatic differentiation is a key feature of present deep learning frameworks. Moreover, they typically provide various ways to specify custom gradients within the computation graph, which is of particular importance for defining surrogate gradients in the realms of non-differentiable operations such as the Heaviside function in spiking neural networks (SNNs). PyTorch, for example, allows the custom specification of the backward pass of an operation by overriding its backward method. Other frameworks provide comparable options. While these methods are common practice and usually work well, they also have several disadvantages such as limited flexibility, additional source code overhead, poor usability, or a potentially strong negative impact on the effectiveness of automatic model optimization procedures. In this paper, an alternative way to formulate surrogate gradients is presented, namely, forward gradient injection (FGI). FGI applies a simple but effective combination of basic standard operations to inject an arbitrary gradient shape into the computational graph directly within the forward pass. It is demonstrated that using FGI is straightforward and convenient. Moreover, it is shown that FGI can significantly increase the model performance in comparison to custom backward methods in SNNs when using TorchScript. These results are complemented with a general performance study on recurrent SNNs with TorchScript and torch.compile, revealing the potential for a training speedup of more than 7x and an inference speedup of more than 16x in comparison with pure PyTorch.
\end{abstract}

\def\sg{\operatorname{sg}}
\newcommand{\sgop}[1]{\sg\mathopen{}\left({#1}\right)\mathclose{}}
\newcommand{\eqd}{=_{\text{\tiny{def}}}}

\renewcommand{\sectionautorefname}{Section}
\renewcommand{\subsectionautorefname}{Section}

\section{Introduction}

Automatic differentiation (or just autograd) is a major conceptual feature in computation graph-oriented deep learning frameworks and thus plays an essential role in the success story of modern artificial intelligence research \citep{Goodfellow2016deep,Paszke2019pytorch,bishop2023learning}. It allows rapid prototyping of complex neural network architectures without the need to specify the gradient calculation by hand. Thus, the backward pass is already implicitly given, when formulating the forward operations. 

There are cases, however, where the backward pass still needs to be specified in a custom manner. This holds true in particular, when forward operations are not differentiable, but a gradient has to be calculated nonetheless. A prominent example is the Heaviside function in spiking neural networks (SNNs). Although this function is not differentiable, a surrogate gradient is defined to enable gradient-based learning \citep{bellec2018long}. Over the recent years, using surrogate gradients, exploring surrogate gradient functions, and even learning them has become a field of research in its own right \citep{neftci2019,li2021differentiable,yin2021accurate,stewart2022meta}

Deep learning frameworks usually provide multiple ways to formulate custom gradients. The gold standard way in PyTorch \citep{Paszke2019pytorch}, for instance, is to override the \texttt{backward()} method of a model or respective submodule as examplarily shown in \autoref{lst:dblgaussiongrad}. Also, various SNN-related frameworks such as SpikingJelly \citep{fang2023spikingjelly}, Norse \citep{pehle2021norse}, or Spyx \citep{kade_heckel_2023_8241588}) provide similar options.

While this undergoing typically fulfills what is needed, it comes with the price of a significant amount of code overhead. If a custom gradient implementation is required on-the-fly this might disrupt the workflow, affect code readability and compatibility, and complicate model prototyping. Moreover, this may also block framework specific builtin optimization routines, such as TorchScript, which is demonstrated in \autoref{sec:results}.

As a possible alternative, this paper introduces \emph{forward gradient injection} (FGI)---a method for custom gradient modeling directly within the forward pass that builds entirely on basic standard operations, without the need to rely on framework specific gradient customization functionality such as overriding the \texttt{backward()} method. FGI is simple, easy-to-use, yet quite effective. Moreover, it can significantly increase the model performance in comparison to custom backward methods in spiking neural networks when using TorchScript.

As a supplementary study, this paper also quantifies the general advantage of applying recent model optimization methods that come with PyTorch, namely, TorchScript and torch.compile in comparison with pure PyTorch models.

\section{Forward Gradient Injection}
Essentially, FGI involves two techniques. The first technique is what is here referred to as \emph{gradient bypassing}. It allows to route the gradient through a different operation as the one used in the forward pass. The second technique is the actual \emph{injection step}, which makes it possible to directly ``inject'' the shape of a local derivative chained in the gradient calculation.

Note that for better readability, some of the code fragments shown to explain FGI are partially incomplete. Further functions and classes can be found separately in \autoref{sec:additionalcode}. Also note that all code listings in this section are based on PyTorch. The principle of FGI, however, should work analogously in all frameworks that provide autograd and the stop gradient operator.

\subsection{Stop Gradient Operator}
An important tool for FGI is the \emph{stop gradient operator} (cf., e.g. the \texttt{detach()} method in PyTorch). For some variable (or computation node) $u$, the stop gradient operator may be formulated as follows.  In forward direction it is just the identity:
\begin{align}
    \sgop{u} \eqd \operatorname{id}(u)
\end{align}
In backward direction, we enforce the following derivative to be zero:
\begin{align}
    \frac{\partial \sgop{u }}{\partial u} \eqd 0
\end{align}
As a result, $u$ will not receive any gradient, since it will always be canceled out (due to multiplication with zero). Note that within the formalism of differential calculus such a construction is, strictly speaking, not valid---it serves us well here nonetheless to formalize the principles behind FGI. It may also be a starting point for a calculus for surrogate gradients (as future research).

\subsection{Gradient Bypassing}
Let $x$ be a variable of interest (node in the computational graph) and let $f(x)$ be an operation for which we want to bypass the gradient via another operation $g(x)$. Gradient bypassing can be realized with the following equation:
\begin{align}
    y = g(x) - \sgop{g(x)} + \sgop{f(x)}
\end{align}

For the forward pass, the above formulation effectively results in $y = f(x)$, since the first two terms cancel themselves out. The same would happen to the gradient received by $x$ in case of $g(x) - g(x)$. But because the second $g(x)$ term---so as $f(x)$---is wrapped with stop gradient, $x$ will receive its gradient from the first $g(x)$ term exclusively:
\begin{align}
\nonumber \frac{\partial y}{\partial x} &= \frac{\partial}{\partial x} \left( g(x) - \sgop{g(x)} + \sgop{f(x)} \right)\\
\nonumber &= \frac{\partial g(x)}{\partial x} - 
\frac{\partial \sgop{g(x)}}{\partial g(x)}\frac{\partial g(x)}{\partial x}
+
\frac{\partial \sgop{f(x)}}{\partial f(x)}\frac{\partial f(x)}{\partial x}\\
&= \frac{\partial g(x)}{\partial x}
\end{align}

As this equation suggests, bypassing allows us to substitute the gradient received by $x$ (or define it in the first place, in case $f(x)$ is not differentiable) with the gradient passed through another operation $g(x)$. \autoref{lst:bypassing} shows how this can be implemented with PyTorch.
\begin{lstlisting}[
    caption=Gradient bypassing,
    label=lst:bypassing
]
def bypass(
    fx: torch.Tensor,
    gx: torch.Tensor
) -> torch.Tensor:
    return gx - gx.detach() + fx.detach()
\end{lstlisting}

This bypassing principle\footnote{Note that what is here referred to a gradient bypassing can be seen as an extension of the straight-through estimator used in JAX \citep{jax2018github}.} calculates the derivative of $g(x)$---via autodifferentiation---and multiplies it into the gradient calculation chain. But it allows us to shape the gradient received by $x$ only indirectly, that is, we have to formulate the antiderivative of the desired derivative shape.

\subsubsection*{\textbf{Example}}

A common way of implementing spiking neurons is applying the Heaviside step function (see \autoref{lst:step}) in forward direction and combining it with surrogate function for the backward pass, since the step function is not differentiable. In PyTorch this is typically realized by creating a new class in which the \lstinline{backward()} method is overridden and customized (cf. \autoref{lst:dblgaussiongrad}).

Let us assume we want to induce the function $\tanh'(x)$ as the surrogate gradient function, because it peaks with $1$ and has a Gaussian-like shape. This can simply be accomplished with the following line of code:
\begin{lstlisting}
y = bypass(step(x), torch.tanh(x))
\end{lstlisting}
Again note that we model the surrogate gradient function in form of its antiderivative, which is transformed into the desired shape by autograd. The left diagram in \autoref{fig:bypass_fgi} shows the output $y$ as well as the gradient of $y$ that is received by $x$ for a range of input values.

\subsection{Gradient Injection via Multiplication}
While gradient bypassing is straightforward to use, especially for simple cases, we may also want to model the surrogate function directly. If the desired shape is more complex, it can be very inconvenient and even inefficient to model the respective antiderivative and let autograd differentiate it. Consider, for instance, a double Gaussian function with negative sections \citep{yin2021accurate} as shown in \autoref{lst:dblgaussian}. The antiderivative of such a function is unnecessarily cumbersome in terms of modeling and computing. In the following, we show that we can model 
the shape of the surrogate derivative directly with a multiplication trick and how we can effectively combine it with bypassing.

The idea of multiplicative gradient injection exploits a simple property of multiplication under differentiation. Let us assume this simple construction: 
\begin{align}
    q = u \cdot v
\end{align}
where $q$ contributes to another variable, e.g., an error term $E$. Deriving $E$ with respect to $u$ unfolds into:
\begin{align}
\frac{\partial E}{\partial u} = \frac{\partial E}{\partial q}\frac{\partial q}{\partial u}
\end{align}
where the last fraction further expands into:
\begin{align}
\frac{\partial q}{\partial u} = \frac{\partial uv}{\partial u} = \frac{\partial u}{\partial u}v + u\frac{\partial v}{\partial u}
\end{align}
In case $v$ is independent of $u$, it holds
\begin{align}\label{eq:dvdu}
\frac{\partial v}{\partial u} = 0
\end{align}
and thus
\begin{align}
\frac{\partial q}{\partial u}=v
\end{align}

Effectively, the gradient that is received by $u$ has exactly the shape of $v$. In case that $v$ depends on $u$ ie. $v(u)$, we can accomplish the same as in \eqref{eq:dvdu} by simply detaching the gradient with $v(\sgop{u})$ or $\sgop{v(u)}$.

We can exploit this multiplication trick and integrate it with the bypassing scheme into the full concept of FGI. Let again $x$ be a tensor of interest, $f(x)$ an operation for which we want to substitute the gradient, and let $g'(x)$ be the shape of the desired surrogate derivative. We formulate FGI through:
\begin{align}
    h &= x \cdot \sgop{g'(x)}\\
    y &= h - \sgop{h} + \sgop{f(x)}
\end{align}

The forward pass will again produce $y=f(x)$ due to out-canceling. When we now compute the derivative of $y$ with respect to $x$ in the backward pass we obtain:

\begin{align}
\frac{\partial y}{\partial x} 
\nonumber &= \frac{\partial}{\partial x} \left(h - \sgop{h} + \sgop{f(x)}\right) \\
\nonumber &= \frac{\partial}{\partial x}\left(x \cdot \sgop{g'(x)}\right) - 
    \frac{\partial \sgop{h}}{\partial h}\frac{\partial h}{\partial x} +
    \frac{\partial \sgop{f(x)}}{\partial f(x)}\frac{\partial f(x)}{\partial x} \\
    & = g'(x)
\end{align}
The PyTorch implementation of FGI wrapped in a function is shown in \autoref{lst:fgi}

\begin{lstlisting}[
    caption=Forward Gradient Injection,
    label=lst:fgi
]
def inject(
    x: torch.Tensor, 
    fx: torch.Tensor, 
    gdx: torch.Tensor
) -> torch.Tensor:
    mul = x * gdx.detach()
    return mul - mul.detach() + fx.detach()
\end{lstlisting}

\subsubsection*{\textbf{Example}}
\begin{figure}[t!]
    \centering
    \includegraphics[width=0.49\linewidth]{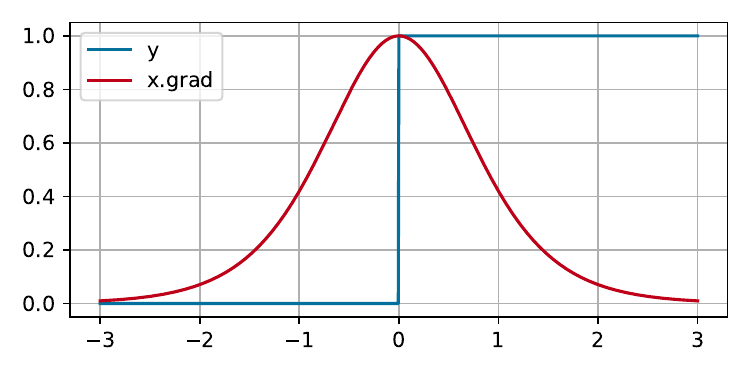}
    \includegraphics[width=0.49\linewidth]{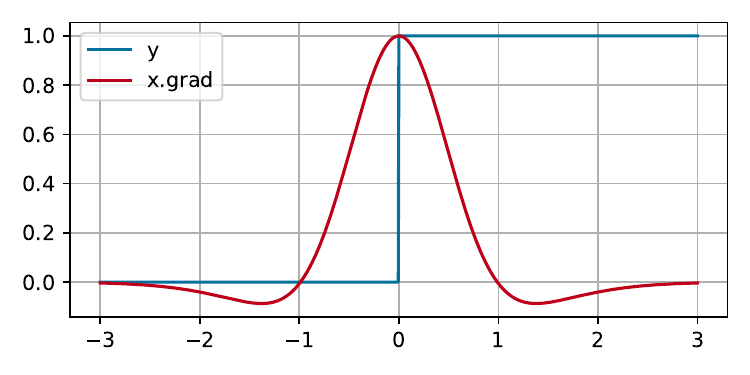}
    \caption{Left: Gradient bypassing example. In forward direction \lstinline{y} is evaluated to \lstinline{step(x)}, whereas in backward direction the gradient is shaped with the derivative of \lstinline{tanh(x)}. Right: Gradient injection example. In forward direction $y$ is again evaluated to $step(x)$, whereas in backward direction the gradient is modulated with $dblgaussian(x)$. In both examples the red curves depict the actual gradient \lstinline{x.grad} retrieved from \lstinline{y.backward()} over a batch input samples.}
    \label{fig:bypass_fgi}
\end{figure}

In this example, we want to implement the above mentioned double Gaussian shape surrogating the non-existent derivative of the step function, analogously to the previous example. We can simply accomplish that with the following function call:

\begin{lstlisting}[]
y = inject(x, step(x), dblgaussian(x))
\end{lstlisting}

Instead of using FGI via a wrapped function, we could also apply it directly inline, reducing additional code dependencies to the minimum:

\begin{lstlisting}
mul = x * dblgaussian(x).detach()
y = mul - mul.detach() + step(x).detach()     
\end{lstlisting}

The right diagram in \autoref{fig:bypass_fgi} shows the output $y$ as well as the gradient that is received by $x$ for a range of input values.

\subsection{Additional PyTorch Functions and Classes}\label{sec:additionalcode}
\begin{lstlisting}[
    caption=Implementation of the Heaviside function,
    label=lst:step
]
def step(x: torch.Tensor) -> torch.Tensor:
    # Alternative:
    # torch.where(x > 0.0, 1.0, 0.0)
    return x.gt(0.0).float()
\end{lstlisting}

\begin{lstlisting}[
    caption=Gaussian function,
    label=lst:gaussian
]
def gaussian(
    x: torch.Tensor, 
    mu: float = 0.0,
    sig: float = 1.0
) -> torch.Tensor:
    return torch.exp(
        -((x - mu) ** 2) / 
        (2.0 * (sig ** 2))
    )
\end{lstlisting}

\begin{lstlisting}[
    caption=Double Gaussian function with negative sections,
    label=lst:dblgaussian
]
def dblgaussian(
    x: torch.Tensor, 
    sig1: float = 0.5,
    sig2: float = 1.0,
    p: float = 0.3
) -> torch.Tensor:
    return (
        (1 + p) * gaussian(x, sig=sig1) -
        p * gaussian(x, sig=sig2)
    )
\end{lstlisting}

\begin{lstlisting}[
    caption=Step function with double gaussian gradient,
    label=lst:dblgaussiongrad
]
class StepDblGaussianGrad(
    torch.autograd.Function
):
    @staticmethod
    def forward(
        ctx, x: torch.Tensor
    ) -> torch.Tensor:
        ctx.save_for_backward(x)
        return step(x)

    @staticmethod
    def backward(
        ctx, grad_output: torch.Tensor
    ) -> torch.Tensor:
        x, = ctx.saved_tensors
        p = 0.3
        dfd = (
            (1 + p) * gaussian(x, sig=0.5) - 
            p * gaussian(x, sig=1.0)
        )
        return grad_output * dfd
\end{lstlisting}

\section{Networks of Spiking Neurons}\label{sec:results}
As a prominent showcase benchmark, we implement FGI in a single hidden layer recurrent network of slightly advanced spiking neurons, namely, adaptive leaky integrate-and-fire (ALIF) neurons \citep{bellec2018long}. ALIF neurons extend on the standard leaky-integrate-and-fire (LIF) spiking neuron by including adaptive threshold dynamics, and have proved to be a suitable compromise in many SNNs between complexity and performance \citep{yin2021accurate,fang2021}.

The ALIF neuron is modeled through the dynamics of two parameters, the membrane potential $u_t$ and the threshold dynamics $\eta_t$, given input $I_t$:
\begin{align}
\label{eq2:adapt}
&u_t = \alpha u_{t-1} + (1-\alpha) I_t - \vartheta_{t-1} z_{t-1}\\
&\eta_t = \gamma \eta_{t-1}+ (1-\gamma)z_{t-1} \\
&\vartheta_t = b_0+\beta\eta_t \\
&z_t = \Theta(u_t - \vartheta_t),
\label{eq2:adaptsn}
\end{align}
where the temporal dynamics are determined by parameters $\alpha,\gamma$, which are computed as $\alpha = \exp(-dt/\tau_m)$ and $\gamma = \exp(-dt/\tau_{adp})$, where $\tau_m$ is the membrane decay time-constant and $\tau_{adp}$ sets the threshold decay. The dynamic threshold is denoted by $\vartheta_t$, having a minimal threshold $b_0$ incremented by scaled adaptive contribution $\beta \eta_t$. $z_t$ is the spiking output of the neuron, which is $0$ when $u_t < \vartheta_t$ and $1$ when $u_t$ exceeds $\vartheta_t$ from below, causing a reset of the neuron. $\Theta$ refers to the non-differentiable Heaviside function, augmented with a surrogate gradient function, for which a single Gaussian function was used in this benchmark.

\subsection{Dataset and Experimental Setup}

The performance impact of FGI is demonstrated using the standard sequential-MNIST dataset (S-MNIST). In the S-MNIST task, the dataset is comprised of the classical MNIST dataset, however, for each sample, pixels are given to the network in a sequential manner: typically 1-by-1, or alternatively k-by-k. Classification in the network is measured at the last timestep, forcing the network to integrate all sequential information in the internal activity of the recurrent network. More specifically, six variants of the problem were used with varying sequence lengths: 28, 49, 98, 196, 392, and 784 time steps---the number of inputs per time step were always 784 divided by the respective sequence length. 

For all sequence lengths, pure Pytorch, TorchScript, and torch.compile were compared for both overriding the \lstinline{backward()} method and FGI. TorchScript and torch.compile are builtin optimization procedures shipped with PyTorch. For every training iteration (with batch size 128) the forward pass and backward pass GPU time was measured separately. Every single experiment comprised of 1,000 training iterations and was repeated ten times. Average values are reported here.

All experiments were pursued with PyTorch 2.2.0 on Python 3.10.13 and CUDA 12.1 (in GPU mode). The host system contained an Intel(R) Core(TM) i7-12700K, 64\,GB DDR4 RAM, and an NVIDIA GeForce RTX 4090 with 24\,GB VRAM.

\subsection{Results and Discussion}

\begin{figure}[t!]

\includegraphics[width=\linewidth]{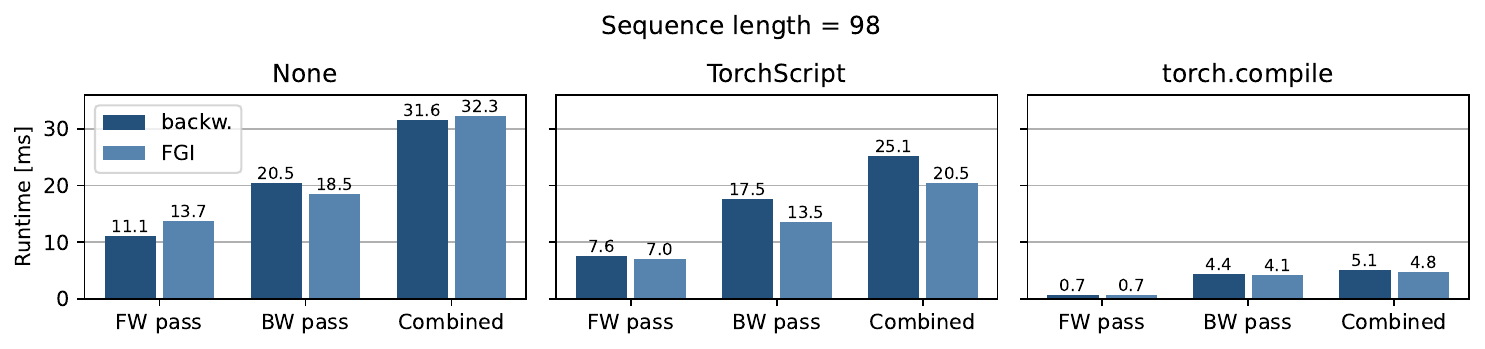}\\
\vspace{0.05cm}
\includegraphics[width=\linewidth]{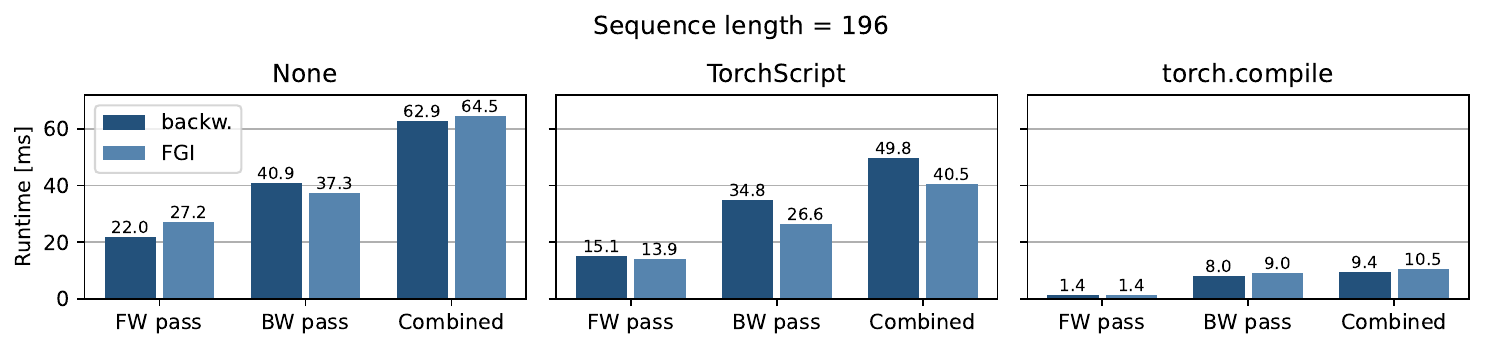}\\
\vspace{0.05cm}
\includegraphics[width=\linewidth]{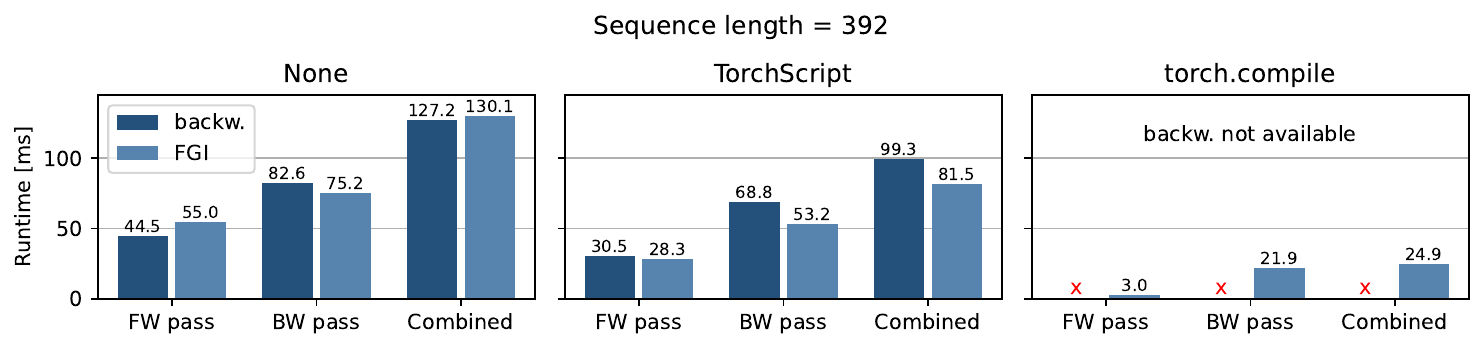}\\
\vspace{0.05cm}
\includegraphics[width=\linewidth]{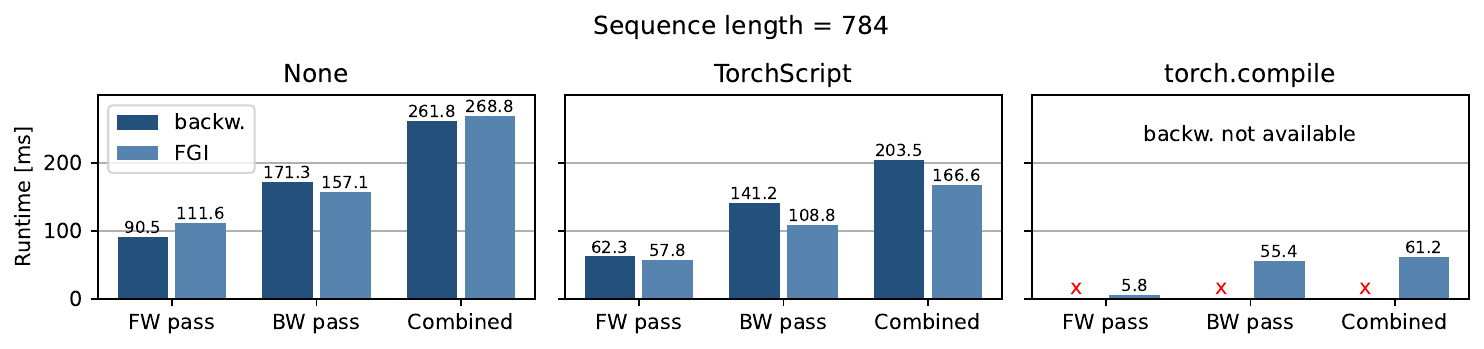}

\caption{Average runtime results (in ms) per batch after warmup for backward-overriding vs. FGI with various sequences lengths contrasting pure PyTorch, TorchScript, and torch.compile. For sequence lengths 392 and 784 torch.compile did not work with custom \lstinline{backward()} implementations but did still work with FGI.}
\label{fig:results}
\end{figure}

The major results are highlighted in \autoref{fig:results} comparing the average runtime for training iteration (split into forward and backward pass). Complete numbers are presented in \autoref{table:training} (for the full training iteration) and in \autoref{table:inference} (for inference only).

When comparing classical custom gradient implementations with FGI, we see that without optimization the forward pass becomes slighlty more costly, where the backwass pass becomes more efficient. In sum there is thus just a marginal difference in runtime. 

TorchScript, however, seems to profit significantly from FGI, as it always improves the runtime. This holds true for the backward pass in particular, indicating that FGI does not obstruct the optimization procedure as much as overriding the \texttt{backward()} method. For the largest sequence length the FGI model is almost twice as fast as the baseline with no optimization. 

On the other hand, with torch.compile there is apparently no significant difference between FGI and the classical approach. However, it is extremely remarkable overall, how well torch.compile accelerates the models. The training is more than seven times faster compared with pure PyTorch (as a reference this is one day instead of a week). When looking at the inference only, the models are even more than 16 times faster (which is more than an order of magnitude).

But this huge acceleration boost comes with a price. The first few iterations with torch.compile in particular, where the models are essentially, dissembled, systematically tested, and compiled into optimized code, can incur significant runtime expenses, as indicated in \autoref{table:warmup}. For the largest sequence length torch.compile requires more than 10 minutes for the first three iterations. Nonetheless, these warmup costs are of course easily amortized in large-scale trainings. One thing to note, however, is that FGI seems to be more favorable for the optimization routines. For the two largest sequence lengtht, torch.compile fails to optimize the models with custom \texttt{backward()} methods, while it can still optimize the FGI models for significant speedups.

It should be highlighted that one clear advantage of TorchScript over torch.compile is that the overhead costs are only marginal in contrast to plain PyTorch (cf. \autoref{table:warmup}). Therefore, one conclusion here is  that for most reactive model exploration, TorchScript seems to be a good option. Moreover, FGI---which profits a lot from TorchScript---could be a powerful tool for rapid model prototyping.

\begin{table}[t!]
\caption{SNN training speedup with model optimization methods over non-optimized baseline.}
\label{table:training}
\small
\begin{tabularx}{\linewidth}{llRRRRRR}
\toprule
\multirow{2}{*}{\textbf{Optimization}} & \multirow{2}{*}{\textbf{Gradient}} & \multicolumn{6}{c}{\textbf{Sequence length}}\\
\cmidrule(lr){3-8}
& & 28 & 49 & 98 & 196 & 392 & 784\\
\midrule
TorchScript & \lstinline{backward()} & 1.2x & 1.2x & 1.3x & 1.3x & 1.3x & 1.3x\\
TorchScript & FGI & 1.5x & 1.5x & 1.5x & 1.6x & 1.6x & 1.6x\\
\rule{0pt}{3ex}%
torch.compile & \lstinline{backward()} & 6.3x & 7.3x & 6.2x & 6.7x & n.a. & n.a\\
torch.compile & FGI & 6.0x & 7.0x & 6.5x & 6.0x & 5.1 &  4.3\\
\bottomrule
\end{tabularx}
\end{table}

\begin{table}[t!]
\caption{SNN inference speedup with model optimization methods over non-optimized baseline.}
\label{table:inference}
\small
\begin{tabularx}{\linewidth}{llRRRRRR}
\toprule
\multirow{2}{*}{\textbf{Optimization}} & \multirow{2}{*}{\textbf{Gradient}} & \multicolumn{6}{c}{\textbf{Sequence length}}\\
\cmidrule(lr){3-8}
& & 28 & 49 & 98 & 196 & 392 & 784\\
\midrule
TorchScript & \lstinline{backward()} & 1.4x & 1.4x & 1.5x & 1.5x & 1.5x & 1.5x\\
TorchScript & FGI & 1.5x & 1.6x & 1.6x & 1.6x & 1.6x & 1.6x\\
\rule{0pt}{3ex}%
torch.compile & \lstinline{backward()} & 12.1x & 14.9x & 16.2x & 15.6x & n.a. & n.a\\
torch.compile & FGI & 12.0x & 14.8x & 16.0x & 15.2x & 15.1x & 15.6x\\
\bottomrule
\end{tabularx}
\end{table}

\begin{table}[t!]
\caption{Warmup costs in seconds summed over the first three training iterations.}
\label{table:warmup}
\small
\begin{tabularx}{\linewidth}{llRRRRRR}
\toprule
\multirow{2}{*}{\textbf{Optimization}} & \multirow{2}{*}{\textbf{Gradient}} & \multicolumn{6}{c}{\textbf{Sequence length}}\\
\cmidrule(lr){3-8}
& & 28 & 49 & 98 & 196 & 392 & 784\\
\midrule
PyTorch & \lstinline{backward()} &  0.6 & 0.7 & 0.7 & 0.8 & 1.1 & 1.4\\
PyTorch & FGI & 0.6 & 0.7 & 0.7 & 0.8 & 1.0 & 1.5\\
\rule{0pt}{3ex}%
TorchScript & \lstinline{backward()} & 1.3 & 1.3 & 1.4 & 1.5 & 1.6 & 2.0\\
TorchScript & FGI & 1.2 & 1.2 & 1.2 & 1.3 & 1.5 & 1.8\\
\rule{0pt}{3ex}%
torch.compile & \lstinline{backward()} & 14.9 & 25.8 & 52.9 & 114.6 & n.a. & n.a\\
torch.compile & FGI & 15.3 & 26.7 & 53.7 & 114.4 & 258.3 &  648.7 \\
\bottomrule
\end{tabularx}
\end{table}

\section{Conclusion}
This study introduced forward gradient injection (FGI) as a novel method for modeling surrogate gradient functions, demonstrating its superiority over traditional custom gradient methods, especially when used with TorchScript. FGI not only simplifies the integration of non-differentiable operations but also significantly enhances model performance and efficiency, here shown in the context of sequence learning with spiking neural networks (SNNs). Moreover, it was shown that torch.compile can immensely accelerate SNN training, reducing the training time by more than 7x and the inference time by more than 16x. 

The simplicity and efficiency of FGI, coupled with its compatibility with TorchScript, highlight its potential as a valuable tool for rapid model prototyping and optimization for handling SNNs. This work underscores the importance of exploring innovative gradient computation techniques to enhance the performance and efficiency of neural network models, particularly in areas where non-differentiable operations are a challenge.

\begin{ack}
Special thanks go to Sander Bohté (CWI, Amsterdam) for enriching and helpful discussions, which inspired and influenced this work a lot. Also many thanks to Saya Higuchi (University of Lübeck) and Co\c{s}ku Can Horuz (University of Lübeck) for reviewing this paper and providing valuable feedback, and to Bernhard Moser (ASAI/SCCH, Austria) for intriguing discussions about the formalism in the context of surrogate gradients.

Sebastian Otte was supported by a Feodor Lynen fellowship of the Alexander von Humboldt Foundation.
\end{ack}

\bibliographystyle{apalike}

\def\bibfont{\small}
\bibliography{literature}

\end{document}